\documentclass{article}
\title{SPGISpeech: 5,000 hours of transcribed financial audio for fully formatted end-to-end speech recognition}
\usepackage{INTERSPEECH2021}
\usepackage{subfigure}
\usepackage{graphics}
\usepackage{graphicx}[realboxes]
\usepackage{eurosym}
\usepackage{rotating}
\usepackage{multicol}
\usepackage{multirow}

\usepackage{xcolor}
\usepackage{url}
\usepackage{hyperref}
\newcommand{\traincalls}{55,289}
\newcommand{\valcalls}{1,114}
\newcommand{\trainslices}{1,966,109}
\newcommand{\valslices}{39,341}
\newcommand{\trainvocab}{100,166}
\newcommand{\valvocab}{19,865}

\newcommand\figref{Fig.~\ref}
\newcommand\tabref{Table~\ref}

\newcommand{\corpus}{SPGISpeech}

\title{\corpus{}: 5,000 hours of transcribed financial audio for fully formatted end-to-end speech recognition}

\name{\begin{tabular}{c}
 Patrick K. O'Neill$^1$,
 Vitaly Lavrukhin$^2$,
 Somshubra Majumdar$^2$,
 Vahid Noroozi$^2$,
 Yuekai Zhang$^3$,\\
 Oleksii Kuchaiev$^2$,
 Jagadeesh Balam$^2$,
 Yuliya Dovzhenko$^1$,
 Keenan Freyberg$^1$,
 Michael D. Shulman$^1$,\\
 Boris Ginsburg$^2$,
 Shinji Watanabe$^{3,4}$,
 Georg Kucsko$^1$
\end{tabular}
}

\address{
  $^1$Kensho Technologies, Cambridge MA, USA\\
  $^2$NVIDIA Technologies, Santa Clara CA, USA\\
  $^3$Johns Hopkins University, Baltimore MD, USA\\
  $^4$Carnegie Mellon University, Pittsburgh PA, USA
}
\email{patrick.oneill@kensho.com, georg@kensho.com}

\begin{document}

\maketitle

\section{Abstract} In the English speech-to-text (STT) machine
learning task, acoustic models are conventionally trained on uncased
Latin characters, and any necessary orthography (such as
capitalization, punctuation, and denormalization of non-standard
words) is imputed by separate post-processing models.
This adds complexity and limits performance, as many formatting tasks
benefit from semantic information present in the acoustic signal but
absent in transcription.  Here we propose a new STT task: end-to-end
neural transcription with fully formatted text for target labels.  We
present baseline Conformer-based models trained on a corpus of 5,000 hours of
professionally transcribed earnings calls, achieving a CER of 1.7.
As a contribution to the STT research community, we release the corpus free for non-commercial use.\footnote{\url{https://datasets.kensho.com/datasets/scribe}}

\section{Introduction}

In the English speech-to-text (STT) task, acoustic model output typically lacks \textit{orthography}, the full set of conventions for expressing the English language in writing (and especially print)\footnote{We concede that the precise details of English orthography can vary by time, geographical region, and even publication house style.  We nevertheless ostensively define \textit{orthographic text} here to mean ``text as it generally appears in U.S. print publications".} \cite{albrow72}. Acoustic models usually render uncased Latin characters, and standard features of English text such as capitalization, punctuation, denormalization of non-standard words, and other formatting information are omitted.  While such output is suitable for certain purposes like
closed captioning, it falls short of the standard of English
orthography expected by readers and can even pose problems for certain
downstream NLP tasks such as natural language understanding, neural
machine translation and summarization \cite{ravichander21, peitz11}.  In contexts where standard
English orthography is required, therefore, it is typically provided
after the fact by a pipeline of post-processing models that infer a
single aspect of formatting each from the acoustic model output.

This approach suffers several drawbacks.  First, rendering orthographically correct text is the more
natural task: if asked to transcribe audio and given no further
instructions, a normally literate English speaker 
will tend to generate orthographic text.  By contrast, writing in block capitals
without punctuation, spelling out numbers in English and so on, is a
far less conventional style.  Secondly, certain types of orthographic
judgments are possible only with acoustic information and cannot be
reliably inferred from text alone. Consider, for example, the problem
of inferring the correct EOS marker for the sentence ``\texttt{the CEO retired}'' (either of a period or question mark) without the information carried in vocal pitch. Third, the practice of chaining orthography-imputing models into pipelines tends to encumber STT systems.  Each orthographic feature may require a model of significant engineering effort in its own right (cf. \cite{sproat16, salloum17, pusateri17, nguyen19, hrinchuk20, sunkara21}), and improvements to one model's performance may degrade end-to-end performance as a whole due to distribution shift \cite{sculley15}.

Modern STT models require large volumes of high-quality training data,
and to our knowledge there is no extant public corpus suitable for the
fully-formatted, end-to-end STT task.  To address this limitation we
release \corpus\footnote{Rhymes with ``squeegee``.}, a subset of the financial audio corpus described
in \cite{huang20}.  \corpus{} offers: 

\begin{itemize}
\item 5,000 hours of audio from real corporate presentations
\item Fully-formatted, professional manual transcription
\item Both spontaneous and narrated speech
\item Varied teleconference recording conditions
\item Approximately 50,000 speakers with a diverse selection of L1 and L2 accents
\item Varied topics relevant to business and finance, including a broad range of named entities.
\end{itemize}

\begin{table*}[ht]
  \centering
  

\begin{tabular}{l|cc}
\hline
  \\ [-2.2ex]
\textbf{Corpus name} & \textbf{Acoustic condition} & \textbf{Speaking style}               \\
\hline
\hline
\\ [-2.2ex]
Switchboard & Telephone          & Spontaneous                                         \\
Librispeech & Close-talk mic.    & Narrated                                             \\
TedLium-3   & Close-talk mic.    & Narrated                                              \\
  Common Voice  & Teleconference     & Narrated       \\
\hline
\\ [-2.2ex]
  \textbf{\corpus{}}  & Teleconference     & Spontaneous, Narrated       \\
  \hline
\end{tabular}
\\
\vspace{10pt}
\begin{tabular}{l|cccc}
\hline
\\ [-2.2ex]
\textbf{Corpus name} & \textbf{Transcription Style} & \textbf{Speaker Count} & \textbf{Vocabulary Size} & \textbf{Amount (h)} \\
\hline
\hline
\\ [-2.2ex]
Switchboard  &Orthographic&550 &25,000 & 300        \\
Librispeech &Non-orthographic &  2,400& 200,000 & 960        \\
TedLium-3   &Non-orthographic &  2,000 &  160,000 & 450        \\
  Common Voice  &Non-orthographic & 40,000 & 220,000 & 1,400\\
\hline
\\ [-2.2ex]
  \textbf{\corpus{}}  & Orthographic &  50,000 & 100,000 & 5,000\\
  \hline
\end{tabular}
  \\ [1.0ex]
    \caption{\textbf{Comparison of \corpus{} to peer corpora.}  For a
  select group of comparable STT corpora, we compare the recording
  format, discursive domain, metadata and quantity of audio.  Numeric data are rounded; precise values for \corpus{} are given in \tabref{tab:stats}.}
\label{tab-comparison}
\end{table*}

\section{Prior Work}
There are many extant STT corpora, varying in their volumes as well as in the details of their formats.
Early corpora include the Wall Street Journal corpus,
consisting of 80 hours of narrated news articles \cite{baker92},
SWITCHBOARD, containing approximately 300 hours of telephone conversations \cite{godfrey92}; TIMIT, consisting of a set of ten
phonetically balanced sentences read by hundreds of speakers ($\sim$
50 h) \cite{garofolo93}; and the Fisher corpus of transcribed
telephone conversations (2,000 h) \cite{cieri04}. 

The LibriSpeech corpus, a standard benchmark for STT models,
consists of approximately 1000 hours of narrated audiobooks
\cite{panayotov15}.  Although the use of audiobooks as an STT
corpus allows researchers to leverage a large pre-existing body of transcription work,
this approach poses several limitations.  First, LibriSpeech
consists entirely of narrated text, hence it lacks many of the
acoustic and prosodic features of spontaneous speech.  Second, as the
narrated books are public domain texts, there is a bias in the corpus
towards older works, and hence against more modern registers of
English.  

Other corpora include
TED-LIUM (450 hours of transcribed TED talks) \cite{hernandez18} and 
Common Voice (a multilingual corpus of narrated prompts with $\sim$ 1,100 validated
hours in English) \cite{ardila19}, and GigaSpeech (10,000 hours from audiobooks, podcasts and YouTube videos) \cite{chen21}.
We present a comparison to select corpora in \tabref{tab-comparison}; a more exhaustive
catalogue of prior STT corpora can be found in \cite{leroux14}.

Within this field of previous work, SPGISpeech is distinctive for being over ten times larger than the next largest corpus with orthographic ground truth labels.  It
also contains approximately 50,000 speakers, the largest number to our knowledge of any public corpus.

\begin{table*}[h]
  \centering
  \scalebox{0.9}{
\begin{tabular}{rl}
  \hline
  \\ [-2.2ex]
  \multicolumn{2}{c}{\textbf{Punctuation}} \\
  \hline
  \\ [-2.2ex]
\textbf{Transcript:} & \texttt{early in April versus what was going on at the beginning of the quarter?
}\\
\textbf{Verbatim:} & \texttt{early you know in april versus uh what was going on at the beginning of the quarter}\\
\textbf{Model Output:}  & \texttt{early in April versus what was going on at the beginning of the quarter?}\\
\\ [-2.2ex]
  \hline
  \\ [-1.5ex]
  \hline
  \\ [-2.2ex]
\multicolumn{2}{c}{\textbf{Non-standard words}} \\
  \hline
  \\ [-2.2ex]
\textbf{Transcript:} & \texttt{[...] for the first time in our 92-year history, we [...]}\\
\textbf{Verbatim:} & \texttt{[...] for the first time in our ninety two year history we [...]}\\
\textbf{Model Output:}  & \texttt{[...] for the first time in our 92-year history, we [...]} \\
  \\ [-2.2ex]
   \hline
   \\ [-1.5ex]
   \hline
  \\ [-2.2ex]
\multicolumn{2}{c}{\textbf{Disfluency}} \\
  \hline
  \\ [-2.2ex]
\textbf{Transcript:} & \texttt{As respects our use of insurance to put out -- reinsurance to put out [...]
}\\
\textbf{Verbatim:} & \texttt{as as respects our use of insurance to put out lim reinsurance to put out [...]}\\
\textbf{Model:} & \texttt{As respects our use of insurance to put out -- reinsurance to put out [...]} \\
  \hline
  \multicolumn{2}{c}{\textbf{Abbreviations}} \\
  \hline
  \\ [-2.2ex]
\textbf{Transcript:} & \texttt{in '15, and got margins back to that kind of mid-teens level [...]}\\
\textbf{Verbatim:} & \texttt{in fifteen and got margins back to that kind of mid teens level [...]}\\
\textbf{Model Output:}  & \texttt{in '15 and got margins back to kind of that mid-teens level [...]} \\
  \\ [-2.2ex]
   \hline
   \\ [-1.5ex]
   \hline
  \\ [-2.2ex]
\end{tabular}
}
 \\ [0.5ex]
  \caption{\textbf{Examples of non-standard text in \corpus.}  For each textual feature we give an example of the ground truth transcript label, a verbatim transcription without casing or punctuation, and Conformer model output with end-to-end orthographic training.}

\label{tab:specialforms}
\end{table*}

\section{Corpus Definition}\label{sec:corpus}

The \corpus{} corpus is derived from company earnings calls
manually transcribed by S\&P Global, Inc. according to a professional style
guide detailing conventions for capitalization, punctuation, denormalization of non-standard words and transcription of disfluencies in spontaneous speech.  The basic unit of \corpus{} is a pair consisting of a 5 to 15 second long 16 bit, 16kHz mono \texttt{wav} audio file and its transcription.

\subsection{Alignment and Slicing}
Earnings calls last 30-60 minutes in length and are typically transcribed as whole units, without internal timestamps.
  In order to produce short audio slices suitable for STT training, the files were segmented with \texttt{Gentle} \cite{gentle20}, a
double-pass forced aligner, with the
beginning and end of each slice of audio imputed by voice
activity detection with \texttt{py-webrtc} \cite{wiseman16}.  While there is inevitably a potential to
introduce certain systematic biases through this process, the fraction
of recovered aligned audio per call ranges from approximately 40\% in
the calls of lowest audio quality to approximately 70\% in the
highest.

\subsection{Corpus Definition}

Slices in \corpus{} are not a simple random sample of the available data, but are subject to certain exclusion criteria.
\begin{enumerate}
\item We sampled no more than four consecutive slices from any call.
We also redacted the corpus out of concern for individual privacy.
Though earnings calls are public, we nevertheless identified full
names with the spaCy \texttt{en\_core\_web\_large} model
\cite{honnibal17}, which we selected on grounds of its wall clock
performance for scanning the entire corpus.  We withheld slices
containing names that appeared fewer than ten times (7\% of total).
Full names appearing ten times or more in the data were considered to
be public figures and were retained.  This necessarily incomplete
approach to named entity recognition was complemented with randomized
manual spot checks which uncovered no false negatives missed by the
automated approach.

\item We excluded all slices that contain currency information (8\% of
total), on the grounds that currency utterances often have non-trivial
denormalizations and misquotation issues that require global context
in order to render correctly.  The utterance \texttt{``one twenty
  three''}, for example, might be correctly transcribed as any of
\texttt{\$1.23}, \texttt{\pounds 1.23}, \texttt{\$1.23 million}, and
so on, depending on context.  Given the potential for material errors
in a business setting if reported incorrectly, moreover, misquotation of
currency values in spontaneous speech are typically corrected in
transcription.  Lacking the means to verify the correct spoken form
for each currency mention, we simply exclude them.

\item We excluded slices with transcriptions containing non-ASCII
characters.  
In particular we excluded all slices whose
transcripts did not consist entirely of the upper- and lowercase ASCII
alphabet; digits; the comma, period, apostrophe, hyphen, question
mark, percent sign, and space characters.

\item Remaining slices were randomly subsampled in order to construct
the published corpus, which consists of two splits, \texttt{train}
and \texttt{val}, having no call events in common.  We also construct a private \texttt{test} split, defined exactly as \texttt{val} and having no calls in common with it, which we do not release.
\end{enumerate}

\section{Corpus Analysis}\label{sec:analysis}

\begin{table}\centering
\begin{tabular}{l|cc}
\hline
  \\ [-2.2ex]
  & \textbf{Train} & \textbf{Val}  \\
  \hline
  \hline
  \\ [-2.0ex]
  Events & \traincalls & \valcalls  \\
  Slices & \trainslices & \valslices \\
  Time (h) & 5,000 & 100\\
  Vocabulary & \trainvocab & \valvocab \\
  OOV & --- &  703\\
\hline
\end{tabular}
 \\ [0.5ex]
\caption{\textbf{\corpus{} Summary Statistics.}}
\label{tab:stats}
\end{table}

We briefly characterize the data.  Summary statistics are given in \tabref{tab:stats}.  Several representative
examples are given in \tabref{tab:specialforms}, highlighting some of the challenges \corpus{} presents for end-to-end training.  The table contains samples from the validation split where the correct EOS marker is difficult to infer from the transcription of the slice alone, non-standard words that must be denormalized, disfluencies and hesitations arising from self-correction in spontaneous speech, and instances like year abbreviations where semantic information is likely necessary to determine the correct orthography.  In each instance we also report characteristic Conformer model output (see Section \ref{sec:model}), demonstrating the feasibility of end-to-end orthographic transcription for these examples.

\corpus{} contains many specialized forms and entity types such as acronyms (15\% of all slices), pauses (10\%), organizations (25\%), persons (8\%), and locations (8\%).  For each form we estimate its prevalence from a random sample of transcripts.
Prevalences of the named entity types \textit{person},
\textit{organization} and \textit{location} were estimated using Flair \cite{akbik18}, which we selected on grounds of its accuracy and acceptable wall clock performance for scanning a small sample of the corpus.  
\begin{figure}[h]
  \centering
  \includegraphics[width=0.45\textwidth]{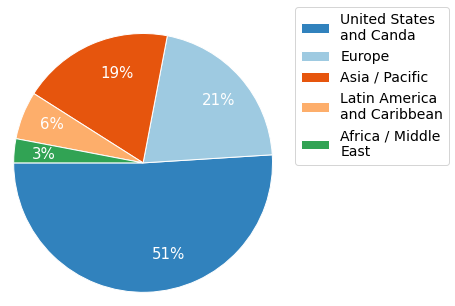}
  \caption{\textbf{Distribution of Speakers by Global Region.}
    Speaker distribution is estimated in a random sample according to
    reported region of corporate domicile.}

  \label{fig:regions}
\end{figure}

There are roughly 50,000 speakers in \corpus{}, drawn from corporate officers and analysts
appearing in English earnings calls and spanning a broad cross-section of
L1 and L2 accents.  We attempted to characterize this linguistic
diversity by tabulating the domiciles of the corporate headquarters of
the companies in the corpus.  In \figref{fig:regions} 
we report the distribution of the global region associated with each
company's headquarters as listed in in the S\&P Capital IQ database \cite{capiq12}.  To further refine the picture of accent
composition, we next considered the distribution of countries of
corporate domicile.  We found it to be long-tailed, the top five
most common nations (US, Japan, UK, India, Canada) comprising
2/3\textsuperscript{rds} of the data.  Although a few of the twenty
most frequent countries are reputed for encouraging strategic
corporate domiciling, we find that their combined share of the total
is low, suggesting that their inclusion does not significantly distort
the bulk statistics.  While imputation of speaker accent from
corporate domicile is a necessarily limited approach, we found the
observed statistics to be broadly concordant with direct estimation of
accent in a random sample.  The number of speakers and diversity of accents in \corpus{} is, to our knowledge, the widest of any public corpus of spontaneous speech, making it especially suitable for training robust STT models.

Lastly, we analyzed the industrial composition of included companies
in order to ensure that they constituted a representative
cross-section of business topics.  All eleven top-level sectors of the Global
Industry Classification Standard (GICS) \cite{gics19} are reflected in
the data at rates roughly comparable to their proportions in the US
economy.  SPGISpeech, not being constrained to any particuar company or industrial sector, therefore offers a fairly synoptic view of modern English business discourse.

\section{Transcription Experiments}\label{sec:model}
To illustrate the feasibility of the end-to-end orthographic approach we train several models on SPGISpeech, including Conformer (ESPnet) and Conformer-CTC (Nemo).  

\subsection{Model Descriptions and Methods}

Both presented models are based on the recently proposed Conformer (Convolution-Augmented Transformer) architecture \cite{gulati2020conformer} which combines the local sensitivities of convolutional neural networks \cite{fukushima80, abdelhamid14}, with the long-range interactions of transformers \cite{vaswani17} in order to capture both local and global dependencies in audio sequences. The unit of prediction for both models consists of SentencePiece tokenized text \cite{richardson18}, whose detokenized output is upper- and lower-case characters, digits, the comma, period, apostrophe, hyphen, question mark, percent sign, and the space character, covering the full range of characters present in \corpus{}.

The ESPnet Conformer model presented consists of 12 conformer blocks with an output dimension of 512 and a kernel size of 31 in the encoder, and 6 transformer blocks in the decoder.
Both encoder and decoder have 8 attention heads with 2048 feed-forward unit dimension. The size of the vocabulary was chosen at $\sim$5k. The model was trained using four 24Gb memory Titan RTX GPUs for 35 epochs. Adam optimizer with no weight decay was used and Noam learning rate scheduler was applied with 25k warmup steps and a learning rate of 0.0015. SpecAug was used with 2 frequency masks and 5 time masks. The last 10 best checkpoints were averaged for the final model. Detailed setups can be found in the ESPnet recipe\footnote{\url{https://github.com/espnet/espnet/tree/master/egs2/spgispeech}\label{foot:espnet}}.

The Conformer-CTC model, provided through the NeMo toolkit \cite{kuchaiev19}, is a CTC-based variant of the Conformer which has the same encoder but uses CTC loss \cite{graves06} instead of RNNT \cite{graves12}. It also replaces the LSTM decoder with a linear decoder on the top of the encoder. This makes
Conformer-CTC a non-autoregressive model unlike the original
Conformer, allowing for significantly faster inference speeds. The presented model was trained for 230 epochs at an effective batch size of 4096, with Adam optimizer and no weight decay. SpecAug was applied with 2 frequency maskings of 27, and 5 time maskings at a maximum ratio of 0.05. Noam learning scheduler was used with a warmup of 10k steps and learning rate of 2.0.

\subsection{Results and Discussion}

As performance on the orthographic STT task hinges in large part on single-character distinctions such as capitalization and punctuation, we report both WER and CER for the \texttt{test} split of SPGISpeech, closely tracking the performance on the \texttt{val} split. We also estimate the respective error rates for a normalized transcription task by lowercasing the output and filtering all but letters, the apostrophe and the space character to conform to the conventional choice of STT vocabulary. Both results are shown in \tabref{tab:wer}.

\begin{table}\centering
\begin{tabular}{l|cccc}
\hline
  \\ [-2.2ex]
  & \multicolumn{2}{c}{\textbf{WER (\%)}} & \multicolumn{2}{c}{\textbf{CER (\%)}} \\
  & \textbf{ortho} & \textbf{norm} & \textbf{ortho} & \textbf{norm} \\
  \hline
  \hline
  \\ [-2.0ex]
  Conformer & \multirow{2}{*}{5.7} & \multirow{2}{*}{2.3} & \multirow{2}{*}{1.7} & \multirow{2}{*}{1.0} \\
  (ESPnet) & & \\
  \\ [-2.0ex]
  \hline
  \\ [-2.0ex]
  Conformer-CTC & \multirow{2}{*}{6.0} & \multirow{2}{*}{2.6}  & \multirow{2}{*}{1.8} & \multirow{2}{*}{1.1} \\
  (NeMo) & & \\
\hline
\end{tabular}
 \\ [0.5ex]
\caption{\textbf{Model Results.} Greedy word error rate and character error rate on the \corpus{} \texttt{test} set. \texttt{ortho} numbers refer to the unnormalized, fully formatted orthographic output, while \texttt{norm} numbers refer to a lowercased and reduced vocabulary to remove the effects of text formatting.}
\label{tab:wer}
\end{table}

Our results demonstrate CERs in the orthographic STT task comparable to those obtained on standard normalized corpora.  While we caution against laying undue stress upon any one particular comparison made across studies, they suggest on the whole that English orthography is within the grasp of modern acoustic architectures, and hence end-to-end orthographic STT is a feasible task.    Both models achieve WERs of less than 6.0 and CERs less than 2.0, making them broadly comparable to previous results obtained with Conformer in other corpora \cite{gulati2020conformer}.

A comparison of the \texttt{ortho} and \texttt{norm} error rates suggests that 1/2 to 2/3\textsuperscript{rds} of the error is due to orthographic issues.  One must recall, however, that there is a lower bound imposed by irreducible error: in some cases there is genuine disagreement as to whether a pause merits a comma, or whether a hesitation should be recorded or emended.  For the disfluency given in \tabref{tab:specialforms}, for example, the model smoothly emends the first repeated word but records the second correction while omitting the partially pronounced word ``\texttt{limit}": some degree of judgment in the training labels, and hence irreducible error, is inevitable.

\section{Conclusions}\label{sec:conclusion}

In this work we introduced a new end-to-end task of fully formatted speech recognition, in which the acoustic model learns to predict complete English orthography. This approach is both conceptually simpler and can lead to improved accuracy due to acoustic cues only present in the the audio, which are needed for full formatting. We demonstrated the feasibility of this approach by training models on \corpus{}, a corpus uniquely suited for the task of large scale, fully formatted end-to-end transcription in English.  As a contribution to the STT research community, we also offer \corpus{} for free academic use.

\section{Acknowledgements}
The authors wish to thank the S\&P Global Market Intelligence Transcripts team for data collection and annotation; Bhavesh Dayalji, Abhishek Tomar, Richard Neale and Gabriela Pereyra for their project support; and Shea Hudson Kerr and Stacey Steele for helpful legal guidance. 

\bibliography{dataset_bib}

\begin{thebibliography}{10}
\providecommand{\url}[1]{#1}
\csname url@samestyle\endcsname
\providecommand{\newblock}{\relax}
\providecommand{\bibinfo}[2]{#2}
\providecommand{\BIBentrySTDinterwordspacing}{\spaceskip=0pt\relax}
\providecommand{\BIBentryALTinterwordstretchfactor}{4}
\providecommand{\BIBentryALTinterwordspacing}{\spaceskip=\fontdimen2\font plus
\BIBentryALTinterwordstretchfactor\fontdimen3\font minus
  \fontdimen4\font\relax}
\providecommand{\BIBforeignlanguage}[2]{{%
\expandafter\ifx\csname l@#1\endcsname\relax
\typeout{** WARNING: IEEEtran.bst: No hyphenation pattern has been}%
\typeout{** loaded for the language `#1'. Using the pattern for}%
\typeout{** the default language instead.}%
\else
\language=\csname l@#1\endcsname
\fi
#2}}
\providecommand{\BIBdecl}{\relax}
\BIBdecl

\bibitem{albrow72}
K.~H. Albrow, ``The english writing system: Notes towards a description,''
  \emph{Schools Council Program in Linguistics and English Teaching, papers
  series 2}, no.~2, 1972.

\bibitem{ravichander21}
\BIBentryALTinterwordspacing
A.~Ravichander, S.~Dalmia, M.~Ryskina, F.~Metze, E.~Hovy, and A.~W. Black,
  ``{NoiseQA: Challenge Set Evaluation for User-Centric Question Answering},''
  in \emph{Conference of the European Chapter of the Association for
  Computational Linguistics (EACL)}, Online, April 2021. [Online]. Available:
  \url{https://arxiv.org/abs/2102.08345}
\BIBentrySTDinterwordspacing

\bibitem{peitz11}
S.~Peitz, M.~Freitag, A.~Mauser, and H.~Ney, ``Modeling punctuation prediction
  as machine translation,'' in \emph{in Proceedings of the International
  Workshop on Spoken Language Translation (IWSLT}, 2011.

\bibitem{sproat16}
R.~Sproat and N.~Jaitly, ``Rnn approaches to text normalization: A challenge,''
  \emph{ArXiv}, vol. abs/1611.00068, 2016.

\bibitem{salloum17}
\BIBentryALTinterwordspacing
W.~Salloum, G.~Finley, E.~Edwards, M.~Miller, and D.~Suendermann-Oeft, ``Deep
  learning for punctuation restoration in medical reports,'' in
  \emph{{B}io{NLP} 2017}.\hskip 1em plus 0.5em minus 0.4em\relax Vancouver,
  Canada,: Association for Computational Linguistics, Aug. 2017, pp. 159--164.
  [Online]. Available: \url{https://www.aclweb.org/anthology/W17-2319}
\BIBentrySTDinterwordspacing

\bibitem{pusateri17}
E.~Pusateri, B.~R. Ambati, E.~Brooks, O.~Pl{\'a}tek, D.~McAllaster, and
  V.~Nagesha, ``A mostly data-driven approach to inverse text normalization,''
  in \emph{INTERSPEECH}, 2017.

\bibitem{nguyen19}
\BIBentryALTinterwordspacing
B.~Nguyen, V.~B.~H. Nguyen, H.~Nguyen, P.~N. Phuong, T.~Nguyen, Q.~T. Do, and
  L.~C. Mai, ``Fast and accurate capitalization and punctuation for automatic
  speech recognition using transformer and chunk merging,'' in \emph{22nd
  Conference of the Oriental {COCOSDA} International Committee for the
  Co-ordination and Standardisation of Speech Databases and Assessment
  Techniques, {O-COCOSDA} 2019, Cebu, Philippines, October 25-27, 2019}.\hskip
  1em plus 0.5em minus 0.4em\relax {IEEE}, 2019, pp. 1--5. [Online]. Available:
  \url{https://doi.org/10.1109/O-COCOSDA46868.2019.9041202}
\BIBentrySTDinterwordspacing

\bibitem{hrinchuk20}
O.~Hrinchuk, M.~Popova, and B.~Ginsburg, ``Correction of automatic speech
  recognition with transformer sequence-to-sequence model,'' in \emph{IEEE
  International Conference on Acoustics, Speech and Signal Processing
  (ICASSP)}, 05 2020, pp. 7074--7078.

\bibitem{sunkara21}
M.~Sunkara, C.~Shivade, S.~Bodapati, and K.~Kirchhoff, ``Neural inverse text
  normalization,'' 2021.

\bibitem{sculley15}
\BIBentryALTinterwordspacing
D.~Sculley, G.~Holt, D.~Golovin, E.~Davydov, T.~Phillips, D.~Ebner,
  V.~Chaudhary, M.~Young, J.-F. Crespo, and D.~Dennison, ``Hidden technical
  debt in machine learning systems,'' in \emph{Advances in Neural Information
  Processing Systems}, C.~Cortes, N.~Lawrence, D.~Lee, M.~Sugiyama, and
  R.~Garnett, Eds., vol.~28.\hskip 1em plus 0.5em minus 0.4em\relax Curran
  Associates, Inc., 2015. [Online]. Available:
  \url{https://proceedings.neurips.cc/paper/2015/file/86df7dcfd896fcaf2674f757a2463eba-Paper.pdf}
\BIBentrySTDinterwordspacing

\bibitem{huang20}
J.~Huang, O.~Kuchaiev, P.~O'Neill, V.~Lavrukhin, J.~Li, A.~Flores, G.~Kucsko,
  and B.~Ginsburg, ``Cross-language transfer learning, continuous learning, and
  domain adaptation for end-to-end automatic speech recognition,'' in
  \emph{International Conference on Multimedia and Expo}, 2021, to appear.

\bibitem{baker92}
D.~B. Paul and J.~M. Baker, ``The design for the {W}all {S}treet
  {J}ournal-based {CSR} corpus,'' in \emph{Speech and Natural Language:
  Proceedings of a Workshop Held at Harriman, New York, {F}ebruary 23-26,
  1992}, 1992.

\bibitem{godfrey92}
J.~J. Godfrey, E.~C. Holliman, and J.~McDaniel, ``{SWITCHBOARD}: Telephone
  speech corpus for research and development,'' in \emph{ICASSP}, 1992.

\bibitem{garofolo93}
J.~S. Garofolo, L.~F. Lamel, W.~M. Fisher, J.~G. Fiscus, D.~S. Pallett, and
  N.~L. Dahlgren, ``{DARPA} {TIMIT} acoustic phonetic continuous speech
  corpus,'' 1993.

\bibitem{cieri04}
\BIBentryALTinterwordspacing
C.~Cieri, D.~Miller, and K.~Walker, ``The {Fisher} {Corpus}: a resource for the
  next generations of speech-to-text,'' in \emph{Proceedings of the Fourth
  International Conference on Language Resources and Evaluation
  ({LREC}{'}04)}.\hskip 1em plus 0.5em minus 0.4em\relax Lisbon, Portugal:
  European Language Resources Association (ELRA), May 2004. [Online].
  Available: \url{http://www.lrec-conf.org/proceedings/lrec2004/pdf/767.pdf}
\BIBentrySTDinterwordspacing

\bibitem{panayotov15}
V.~Panayotov, G.~Chen, D.~Povey, and S.~Khudanpur, ``Librispeech: an {ASR}
  corpus based on public domain audio books,'' in \emph{Acoustics, Speech and
  Signal Processing (ICASSP)}, 2015.

\bibitem{hernandez18}
F.~Hernandez, V.~Nguyen, S.~Ghannay, N.~Tomashenko, and Y.~Estève,
  ``{TED-LIUM} 3: twice as much data and corpus repartition for experiments on
  speaker adaptation,'' 2018.

\bibitem{ardila19}
R.~Ardila, M.~Branson, K.~Davis, M.~Henretty, M.~Kohler, J.~Meyer, R.~Morais,
  L.~Saunders, F.~Tyers, and G.~Weber, ``{Common Voice}: A
  massively-multilingual speech corpus,'' 2019.

\bibitem{chen21}
G.~Chen, S.~Chai, G.~Wang, J.~Du, C.~W. Wei-Qiang~Zhang, D.~Su, D.~Povey,
  J.~Trmal, J.~Zhang, M.~Ji, S.~Khudanpur, S.~Watanabe, S.~Zhao, W.~Zou, X.~Li,
  X.~Yao, Y.~Wang, Z.~You, and Z.~Yan, ``{GigaSpeech}: An evolving,
  multi-domain {ASR} corpus with 10,000 hours of transcribed audio,''
  \emph{submitted to INTERSPEECH}, 2021.

\bibitem{leroux14}
J.~Le~Roux and E.~Vincent, ``{A categorization of robust speech processing
  datasets},'' Mitsubishi Electric Research Labs TR2014-116, Technical Report,
  2014.

\bibitem{gentle20}
\BIBentryALTinterwordspacing
lowerquality, \emph{Gentle Aligner}, 2020 (accessed May 4th, 2020). [Online].
  Available: \url{https://lowerquality.com/gentle/}
\BIBentrySTDinterwordspacing

\bibitem{wiseman16}
\BIBentryALTinterwordspacing
J.~Wiseman, \emph{pyWebRTC}, 2016 (accessed May 4th, 2020). [Online].
  Available: \url{https://github.com/wiseman/py-webrtcvad}
\BIBentrySTDinterwordspacing

\bibitem{honnibal17}
M.~Honnibal and I.~Montani, ``{spaCy 2}: Natural language understanding with
  {B}loom embeddings, convolutional neural networks and incremental parsing,''
  2017, to appear.

\bibitem{akbik18}
A.~Akbik, D.~Blythe, and R.~Vollgraf, ``Contextual string embeddings for
  sequence labeling,'' in \emph{{COLING} 2018, 27th International Conference on
  Computational Linguistics}, 2018, pp. 1638--1649.

\bibitem{capiq12}
S.~Global, ``S\&p capital iq,'' \url{www.capitaliq.com}, 2012, accessed June
  2020.

\bibitem{gics19}
\BIBentryALTinterwordspacing
M.~S.~C. International, ``The global industry classification standard
  ({GICS}),'' 2019. [Online]. Available: \url{https://www.msci.com/gics}
\BIBentrySTDinterwordspacing

\bibitem{gulati2020conformer}
A.~Gulati, J.~Qin, C.-C. Chiu, N.~Parmar, Y.~Zhang, J.~Yu, W.~Han, S.~Wang,
  Z.~Zhang, Y.~Wu \emph{et~al.}, ``Conformer: Convolution-augmented transformer
  for speech recognition,'' \emph{arXiv preprint arXiv:2005.08100}, 2020.

\bibitem{fukushima80}
K.~Fukushima, ``{N}eocognitron: {A} self-organizing neural network model for a
  mechanism of pattern recognition unaffected by shift in position,''
  \emph{Biological Cybernetics}, vol.~36, pp. 193--202, 1980.

\bibitem{abdelhamid14}
\BIBentryALTinterwordspacing
O.~Abdel{-}Hamid, A.~Mohamed, H.~Jiang, L.~Deng, G.~Penn, and D.~Yu,
  ``Convolutional neural networks for speech recognition,'' \emph{{IEEE} {ACM}
  Trans. Audio Speech Lang. Process.}, vol.~22, no.~10, pp. 1533--1545, 2014.
  [Online]. Available: \url{https://doi.org/10.1109/TASLP.2014.2339736}
\BIBentrySTDinterwordspacing

\bibitem{vaswani17}
\BIBentryALTinterwordspacing
A.~Vaswani, N.~Shazeer, N.~Parmar, J.~Uszkoreit, L.~Jones, A.~N. Gomez, L.~u.
  Kaiser, and I.~Polosukhin, ``Attention is all you need,'' in \emph{Advances
  in Neural Information Processing Systems}, I.~Guyon, U.~V. Luxburg,
  S.~Bengio, H.~Wallach, R.~Fergus, S.~Vishwanathan, and R.~Garnett, Eds.,
  vol.~30.\hskip 1em plus 0.5em minus 0.4em\relax Curran Associates, Inc.,
  2017. [Online]. Available:
  \url{https://proceedings.neurips.cc/paper/2017/file/3f5ee243547dee91fbd053c1c4a845aa-Paper.pdf}
\BIBentrySTDinterwordspacing

\bibitem{richardson18}
\BIBentryALTinterwordspacing
T.~Kudo and J.~Richardson, ``{S}entence{P}iece: A simple and language
  independent subword tokenizer and detokenizer for neural text processing,''
  in \emph{Proceedings of the 2018 Conference on Empirical Methods in Natural
  Language Processing: System Demonstrations}.\hskip 1em plus 0.5em minus
  0.4em\relax Brussels, Belgium: Association for Computational Linguistics,
  Nov. 2018, pp. 66--71. [Online]. Available:
  \url{https://www.aclweb.org/anthology/D18-2012}
\BIBentrySTDinterwordspacing

\bibitem{kuchaiev19}
O.~Kuchaiev, J.~Li, H.~Nguyen, O.~Hrinchuk, R.~Leary, B.~Ginsburg, S.~Kriman,
  S.~Beliaev, V.~Lavrukhin, J.~Cook, P.~Castonguay, M.~Popova, J.~Huang, and
  J.~M. Cohen, ``Nemo: a toolkit for building ai applications using neural
  modules,'' 2019.

\bibitem{graves06}
A.~Graves, S.~Fernández, and F.~Gomez, ``Connectionist temporal
  classification: Labelling unsegmented sequence data with recurrent neural
  networks,'' in \emph{In Proceedings of the International Conference on
  Machine Learning, ICML 2006}, 2006, pp. 369--376.

\bibitem{graves12}
A.~Graves, ``Sequence transduction with recurrent neural networks,'' in
  \emph{ICML 29}, 2012.

\end{thebibliography}
\bibliographystyle{IEEEtran}

\end{document}